\documentclass[final]{cvpr}

\usepackage{times}
\usepackage{epsfig}
\usepackage{graphicx}
\usepackage{amsmath}
\usepackage{amssymb}
\usepackage{soul}

\usepackage[symbol]{footmisc}

\usepackage[pagebackref=true,breaklinks=true,colorlinks,bookmarks=false]{hyperref}

\def\cvprPaperID{2122} 
\def\confYear{CVPR 2021}

\begin{document}

\title{Progressive Temporal Feature Alignment Network for Video Inpainting}

\author{Xueyan Zou$^{2}$\thanks{Work mostly done during an internship at ByteDance Inc.}


\and
Linjie Yang$^1$
\and
Ding Liu$^1$
\and
Yong Jae Lee$^2$
\and
\centerline{\tt\small  \{zxyzou,yongjaelee\}@ucdavis.edu, \{linjie.yang,liuding\}@bytedance.com}
\and
\centerline{ \textsuperscript{\rm 1}ByteDance Inc. \; \;\;\;\;\;\;  \textsuperscript{\rm 2}University of California, Davis}
}

\maketitle
\begin{abstract}
Video inpainting aims to fill spatio-temporal ''corrupted'' regions with plausible content. To achieve this goal, it is necessary to find correspondences from neighbouring frames to faithfully hallucinate the unknown content. Current methods achieve this goal through attention, flow-based warping, or 3D temporal convolution. However, flow-based warping can create artifacts when optical flow is not accurate, while temporal convolution may suffer from spatial misalignment. We propose `Progressive Temporal Feature Alignment Network', which progressively enriches features extracted from the current frame with the feature warped from neighbouring frames using optical flow. Our approach corrects the spatial misalignment in the temporal feature propagation stage, greatly improving visual quality and temporal consistency of the inpainted videos.
Using the proposed architecture, we achieve state-of-the-art performance on the DAVIS and FVI datasets compared to existing deep learning approaches. Code is available at \url{https://github.com/MaureenZOU/TSAM}.
\end{abstract}


\section{Introduction}
Video inpainting is a task that aims to fill missing regions in video frames with plausible content \cite{bertalmio2001navier}. It has a wide range of applications including corrupted video restoration, watermark/logo removal, object removal, etc. To fill the ``holes'', it is ideal to inpaint them with corresponding known content from neighbouring frames, which can well approximate the missing region. For any missing pixels that lack good correspondence due to e.g., occlusion, the video inpainting method must hallucinate reasonable content. 

Existing state-of-the-art video inpainting methods rely on extracting useful information from neighbouring frames, and are based on three main directions: 3D temporal convolution~\cite{chang2019free, chang2019learnable, huproposal}, optical flow~\cite{gao2020flow, xu2019deep}, and attention~\cite{zeng2020learning, oh2019onion}.

\begin{figure*}[t]
\begin{center}
\includegraphics[width=1.\textwidth]{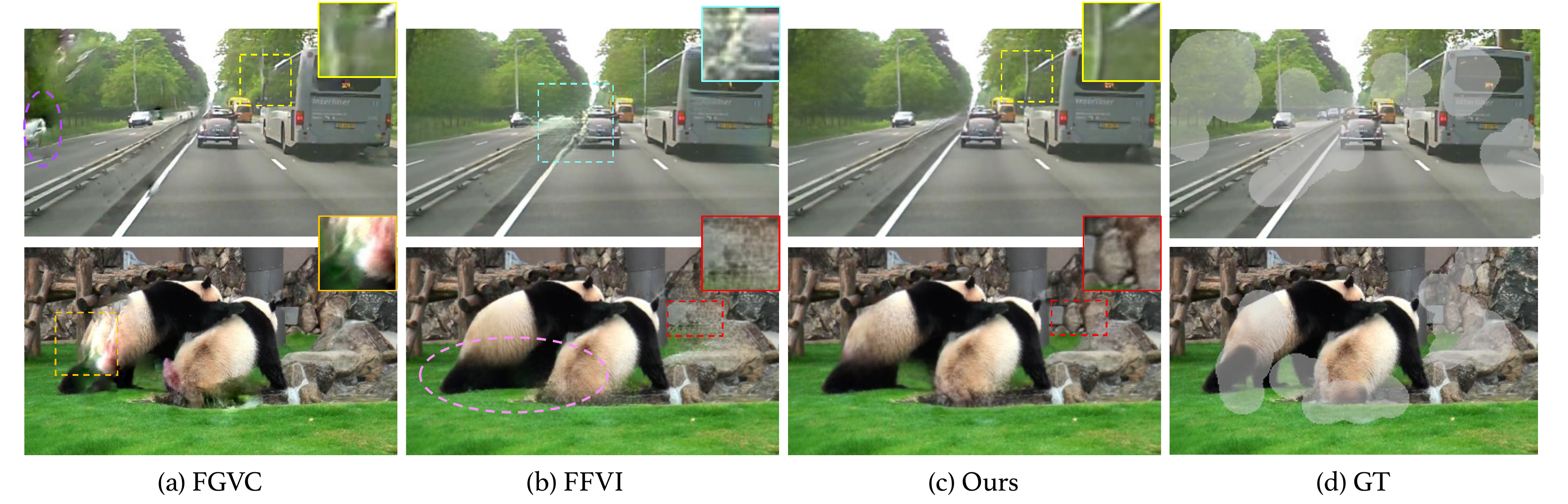}
\end{center}
\vspace{-12pt}
   \caption{This figure shows a qualitative comparison of our approach with flow-based method FGVC~\cite{gao2020flow} and 3D convolution based method FFVI~\cite{chang2019free}. We try to inpaint the gray regions shown in (d). The results show that our method generates content that is both more structure-preserving and visually appealing.}
\label{fig:intro}
\end{figure*}

The general structure of existing 3D convolution models for video inpainting consists of a fully convolutional generator to predict the inpainted result holistically and a Temporal-Patch GAN discriminator to enforce temporal smoothness and frame realism~\cite{chang2019learnable,chang2019free}. However, these methods simply stack feature maps from neighboring frames to encode 3D temporal information as an additional axis, without considering the movement of objects across those frames.  This leads to spatial misalignment in the features, which can cause issues for video inpaiting.  For example, as shown in the pink circle on the second row of Fig. \ref{fig:intro}, the panda leg is not successfully inpainted when using 3D convolutions~\cite{chang2019free}.  In order to accurately predict structural (i.e., edge/shape) details, video inpainting models require spatially-aligned feature maps for each timestamp.  This motivates us to propose a feature alignment framework to overcome these challenges, with inspirations from flow-based approaches. 



Recent flow-based approaches~\cite{gao2020flow, xu2019deep} first compute optical flow from the corrupted video frames, and then complete the unknown regions using flow-based inpainting techniques. Using the computed flow, pixels in corrupted regions are propagated from adjacent frames. Further, image inpainting techniques such as \cite{yu2018generative} are applied to complete the remaining content. Although optical flow methods are good at spatial content alignment and are able to inpaint video frames with higher resolution compared to attention or 3D convolution models, any errors in optical flow (especially in the missing regions) can prevent the models from capturing fine-grained structural details. For example, as shown in Fig.~\ref{fig:intro}, the telephone pole in the first row using FGVC~\cite{gao2020flow}, a state-of-the-art optical flow based approach, is not straight compared to 3D convolution approaches (FFVI~\cite{chang2019free} and ours). In addition, image inpainting techniques might generate unwanted content that does not match the ground truth content. As shown in the first row of Fig. \ref{fig:intro}, the purple circle denotes an area that could not be handled by propagated pixels. FGVC generates a car on the grass that does not exist in the original video.

We hereby propose a novel framework called Progressive Temporal Feature Alignment Network to combine the advantages and offset the weaknesses of temporal convolution frameworks and optical flow based warping approaches. Our method is an end-to-end deep network with a novel temporal shift-and-aligned module (TSAM), in which features between neighbouring frames are aligned using optical flow. In order to extract aligned feature representations across different scales, we progressively apply TSAM to feature maps in different scales at different network depths in a coarse-to-fine manner.  Fig.~\ref{fig:intro} shows that our method produces satisfactory results in challenging cases in terms of spatial alignment, resolution, and coarse to fine-grained structures. Through extensive experiments, we demonstrate that our method achieves state-of-the-art performance on two video inpainting benchmarks FVI~(subset of YoutubeVOS)~\cite{chang2019free} and DAVIS~\cite{Caelles_arXiv_2018}.

\begin{figure*}[t]
\begin{center}
\includegraphics[width=1.\textwidth]{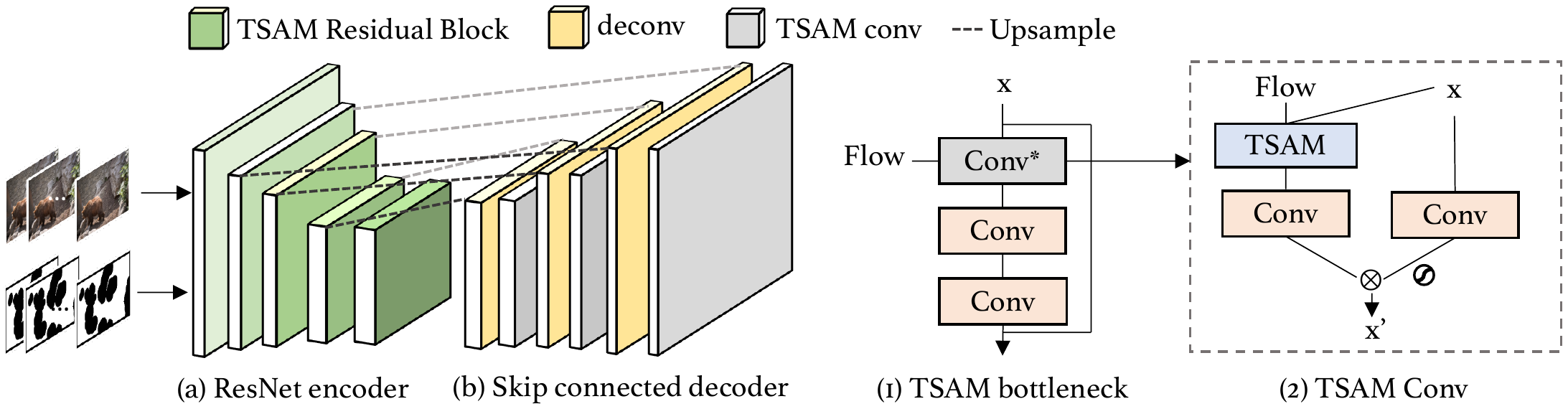}
\end{center}
\vspace{-8pt}
   \caption{
      The generator of our temporal feature alignment network. It consists of (a) a ResNet encoder with all the first Conv layer in the bottleneck block replaced with our TSAM Conv (refer to (1)); and (b) a skip connected decoder that has 3 gated DeConv layers and 5 TSAM Conv layers (only three are shown). As shown in (2), the TSAM Conv layer consists of three parts including a temporal shift-and-align module and 2 conv layers. The temporal feature $x$ is first aligned using optical flow. After going through a conv layer, the temporal feature is passed through a gating signal (via a dot product) which is also computed using $x$ through conv and sigmoid layers. 
      }
\label{fig:pipeline}
\end{figure*}


\section{Related Work}

\vspace{-1pt}
\subsection{Image Inpainting}
Image inpainting aims to inpaint the missing region with retrieved or synthesised content. Traditional methods either retrieve or interpolate missing content from the image itself \cite{ballester2001filling, levin2003learning, barnes2009patchmatch} or a database of related images \cite{hays2007scene}. Recent deep learning approaches achieve better results~\cite{kohler2014mask, ren2015shepard}, in particular, with the advent of Generative Adversarial Networks (GAN) \cite{goodfellow2014generative} which can make the inpainted images more realistic \cite{pathak2016context, iizuka2017globally}. However, GANs can synthesize content that is unrelated to the original image.  Image inpainting algorithms \cite{yu2018generative} have been applied to video inpainting methods \cite{xu2019deep,gao2020flow}. All of the above methods share a same limitation: the missing region is hallucinated from the surrounding region or from external image databases, which can be suboptimal for video which can exploit temporal redundancy from neighboring frames. 


\subsection{Video Inpainting}
In video inpainting, the information of the corrupted region can have the possibility to be retrieved from nearby frames thanks to the temporal consistency of videos. Traditional methods are usually patch-based \cite{granados2012not, huang2016temporally, newson2014video}, which can generate plausible results under certain conditions (e.g. repetitive patterns, similar textures) but often come with a high computational cost. Recent deep learning based video inpainting methods propose more efficient and effective solutions, and include three main directions: attention based mechanisms \cite{oh2019onion, zeng2020learning, li2020short}, flow guided approaches \cite{xu2019deep, gao2020flow}, and 3D convolutional networks~\cite{chang2019free, chang2019learnable, huproposal}. These methods use different techniques to borrow information from neighbouring frames. Attention based methods retrieve information from neighboring frames using a weighted sum that can lead to blurry results. Flow guided approaches are able to generate higher resolution results but are sensitive to errors in optical flow. 3D convolutional networks are efficient with an end-to-end structure, but can suffer from spatial misalignment and lower resolution in the inpainted area.  Combining the ideas of 3D convolutional networks and flow-guided methods, our approach is an end-to-end 3D convolutional framework with an embedded temporal shift-and-align module, which enables accurate temporal feature alignment and propagation.


\subsection{Temporal Modeling}
To handle temporal information, C3D~\cite{tran2014c3d} propose 3D spatio-temporal CNNs. Later, I3D~\cite{carreira2017quo} propose to inflate all the 2D convolution filters into 3D convolutions. In order to improve the time efficiency of 3D convolutions, \cite{tran2018closer, xie2018rethinking} propose to combine 2D and 3D convolution. The Temporal shift module (TSM)~\cite{lin2019tsm} combines 2D convolution and channel shifting across temporal features to mimic 3D convolution, and shows performance gains on action recognition and video object detection as well as improved time efficiency for video inpainting \cite{chang2019learnable}. We regard the TSM network as a type of 3D convolution in this paper as it uses temporal information across frames in its basic blocks (e.g. each bottleneck block in ResNet \cite{he2016deep}). \cite{chang2019learnable} applies TSM in an end-to-end framework for video inpainting. However, directly shifting the features from adjacent frames introduces semantic misalignment on the feature map. Our approach introduces a spatially-aligned version of TSM to fix the misalignment for video inpainting.


\section{Method}\label{section:model}

In this section, we first give an overview of our model design for video inpainting. We then introduce our temporal shift-and-align module, which builds upon the temporal shift module~\cite{lin2019tsm}. Finally, we introduce the loss functions used to train our model. 


\subsection{Overview}
\paragraph{Problem Definition}
Video inpainting can be formulated as a conditional pixel prediction task: given ordered input video frames $X^T = [f_1, f_2, ..., f_T]$ with corrupted regions $M = [m_1, m_2, ..., m_T]$, the objective is to predict the original video $Y^T = [F_1, F_2, ..., F_T]$. Each $m_i$ is a binary mask with the same resolution as the video frames where $0$ indicates the pixel is missing or corrupted and $1$ indicates the pixel is valid.

\begin{figure*}[t]
\begin{center}
\includegraphics[width=1.02\textwidth]{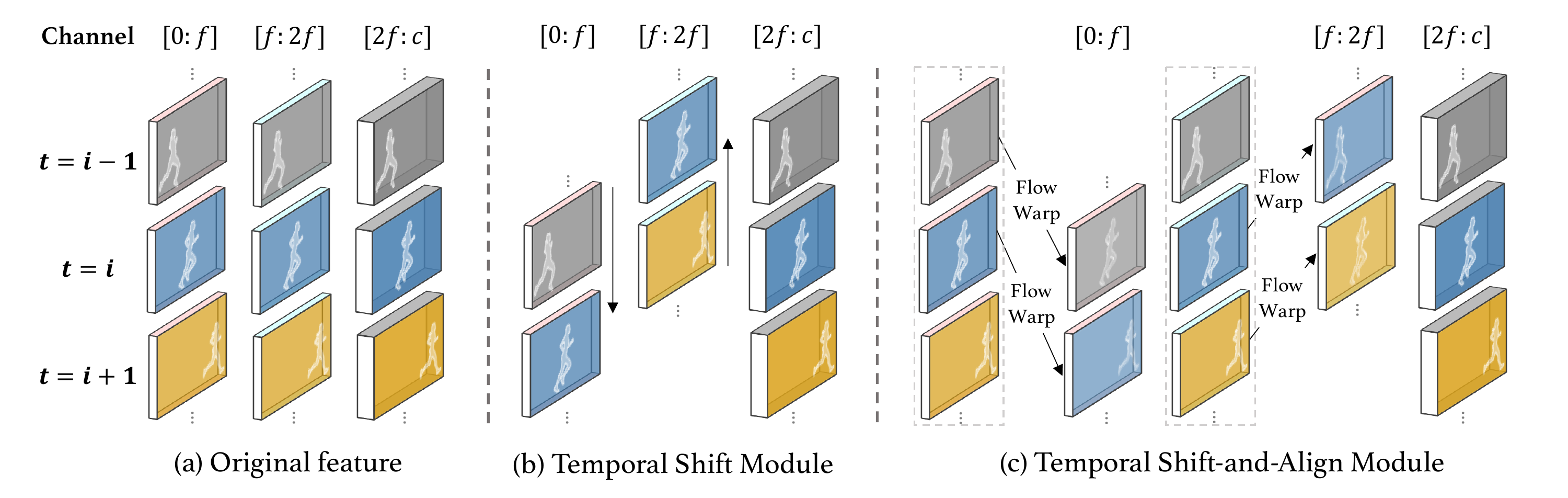}
\end{center}
\vspace{-8pt}
   \caption{
   This figure shows the comparison between the temporal shift module \cite{lin2019tsm} and our temporal shift-and-align module. (a) The original feature, which contains three feature maps at three different timestamps that are split into three channel groups. (b) The temporal shift module shifts the first $[0:f]$ channels downwards and $[f:2f]$ channels upwards. In this way, the feature at time stamp $t=i$ also contains information from $t=i-1$ and $t=i+1$. (c) Our Temporal Shift-and-Align module warps the shifted features from the neighboring frames to be spatially aligned with the features in the current timestamp.}
\label{fig:fgtsm}
\vspace{-8pt}
\end{figure*}

\vspace{-5pt}
\paragraph{Model Design}
As shown in Fig.~\ref{fig:pipeline}, our model consists of three parts: (1) \textit{A ResNet~\cite{he2016deep} encoder backbone}, with the first convolution layer of every bottleneck block replaced with TSAM convolution. TSAM convolution takes both feature maps and optical flow as input.  It first shifts the features of the neighbouring frames, and then uses optical flow to warp the shifted features to the correct spatial location at the current shifted time stamp.  We use gated convolution to mitigate any side effects brought by missing regions. The gating signal is computed using the original feature map through a convolution layer and a sigmoid layer. The final output of TSAM Conv is the dot product between the computed feature and gating signal. (2) \textit{A skip connected decoder} that contains 3 gated deconvolution layers and 5 TSAM convolution layers with gating signal. There are two convolution layers that are used for channel reduction, which are not shown in Fig. \ref{fig:pipeline}. The ResNet encoder and skip connected decoder together make up the generator, which inpaints the corrupted pixels by borrowing information from neighboring frames via 3D convolution \cite{chang2019free, chang2019learnable} and hallucinating any remaining missing content with the help of adversarial loss \cite{goodfellow2014generative}, perceptual loss and etc.. (3) \textit{A temporal patch GAN discriminator} to enforce the spatio-temporal features to follow the ground truth target distribution. 

\subsection{Temporal Shift Module (TSM)}
TSM \cite{lin2019tsm} is a temporal feature shifting method to exchange information between neighboring frames on the channel dimension. It is usually combined with 2D convolution to mimic the effect of 3D convolutions with reduced memory and latency. 
As shown in Fig.~\ref{fig:fgtsm} (b), the channels with index $[0:f]$ are shifted downward, and channels with index $[f:2f]$ are shifted upward, so that the feature map at $t=i$ will be enriched with features from $t=i+1$ and $t=i-1$. Each shift operation introduces a temporal window size of $3$. As the network gets deeper and adopts more TSM modules, the temporal receptive field increases linearly as $2n-1$ with respect to the number of TSM modules $n$ inserted.


Although TSM module efficiently aggregates temporal information, the aggregated temporal features are not spatially-aligned. As shown in Fig.~\ref{fig:fgtsm} (b), the positions of the person are at different locations in different frames due to object motion. 
As a result, the aggregated TSM features will be misaligned in terms of image content, which can cause the inpainted frame to be blurry (Fig.~\ref{fig:intro} (b), cyan square) or spatially-misaligned (Fig.~\ref{fig:intro} (b) pink circle). In order to solve this problem, we propose the Temporal Shift-and-Align module as follows.


\subsection{Temporal Shift-and-Align module (TSAM)}
Our Temporal Shift-And-Align module consists of three steps: (1) Shift the features of neighboring frames. (2) Use optical flow to warp the shifted features to the correct spatial location at the current timestamp. (3) Aggregate the spatially-aligned neighbour features with the current frame features using a validity mask. We describe in detail each step in the following sections.



\begin{figure*}[t]
\begin{center}
\includegraphics[width=1.\textwidth]{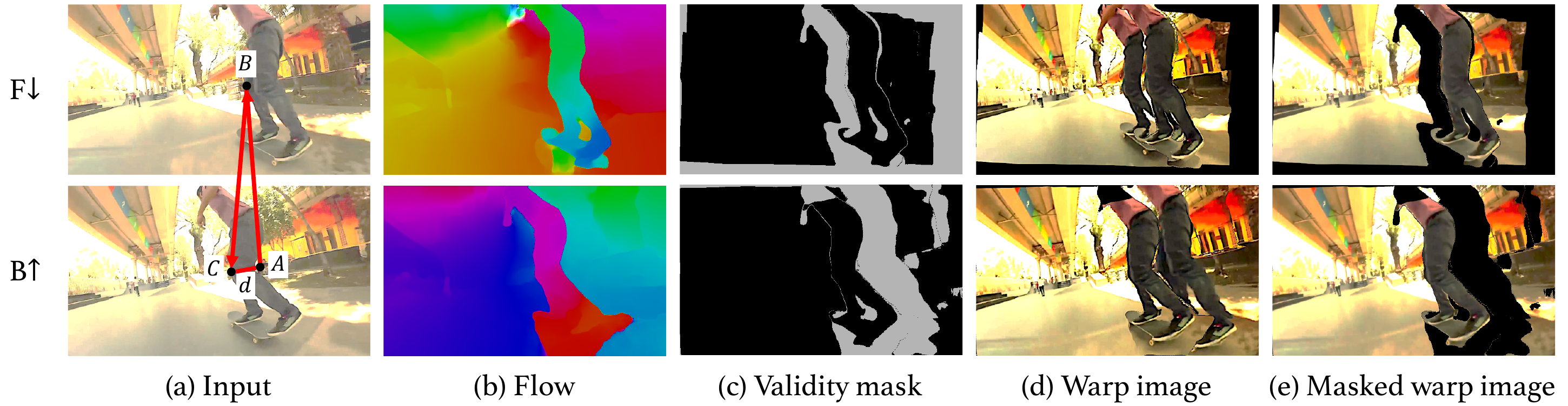}
\end{center}
\vspace{-12pt}
   \caption{This figure shows the optical flow computed on a pair of images. (a) The input image pair. (b) Computed optical flow. The first row is the forward flow and the second row is the backward flow. (c) Computed validity mask for forward and backward flow. (d) Warped image without validity mask. (e) Warped image with validity mask. We can see that if we simply use the reverse forward flow to warp the first image to the second image, an extra person is hallucinated in the warped image (d). This is because the background region to the left of the human does not have corresponding pixels in the second image as it is occluded by the human.}
\label{fig:cycle_flow}
\vspace{-8pt}
\end{figure*}

\vspace{-5pt}
\paragraph{Optical Flow}
Optical flow is defined as the offset between corresponding pixels in a pair of images $(I^{t},I^{t+\Delta t} )$ within a small time interval $\Delta t$.  The offset $(\Delta x, \Delta y)$ generally follows:
\begin{equation}
    I^t(x,y) = I^{t+\Delta t}(x+\Delta x, y+ \Delta y).
\end{equation}
The collection of the offsets on an image is defined as a flow map $F$. Although current optical flow methods \cite{ilg2017flownet, sun2018pwc, teed2020raft} can generate accurate results on real world videos, they still struggle to produce accurate optical flow in difficult cases such as occlusion and fast motion. Fig.~\ref{fig:cycle_flow} (a) shows a pair of images and Fig.~\ref{fig:cycle_flow} (b) shows their forward and backward optical flows. As an example, the first warped images in Fig.~\ref{fig:cycle_flow} (d) are computed using the reverse forward flow to warp the pixels from the second image back to the first image. However, directly warping the image using optical flow generates `ghosting humans' due to occlusion, since some pixels in the first image do not have correspondences in the second image. Thus, if we simply use the warped features for feature aggregation, misaligned pixels will exist. To overcome this issue, we calculate the validity mask to mark the reliable pixels in the flow. 


\vspace{-5pt}
\label{section:agg}
\paragraph{Flow Validity Mask}
We regard the flow at location $(x,y)$ to be invalid in two cases: (1) we cannot find a mapping from the reference image; or (2) the flow calculated is inaccurate due to occlusion, fast motion, etc. We take advantage of the cycle consistency of optical flow (e.g.,~\cite{lai2019self}) to detect the invalid pixels in both cases. As shown in Fig. \ref{fig:cycle_flow} (a), flow computed at location A is valid if and only if:
\begin{equation}
   || A - F_{f}(F_{b}(A)) || < \delta 
\end{equation}
where $F_f$ and $F_b$ are forward and backward flow maps, respectively. A validity mask can be computed for both forward and backward flow due to the cycle structure of optical flow between two images. We show validity masks of the optical flow in Fig.~\ref{fig:cycle_flow} (c) where grey regions denote the invalid pixels calculated under cases (1) and (2). To better understand the validity mask, Fig.~\ref{fig:cycle_flow} (d) show the images warped using  optical flow computed by flownet2~\cite{ilg2017flownet}. By multiplying the validity mask with the warped image, we get the masked warped images in Fig.~\ref{fig:cycle_flow} (e), in which the inaccurate warped pixels have been masked out.

\vspace{-5pt}
\label{section:agg}
\paragraph{Shift-and-Align for Feature Aggregation}

After computing the optical flow maps and the corresponding validity masks, we use them in our TSAM module. As shown in Fig. \ref{fig:fgtsm}~(c), the TSAM module consists of three operations: (1) Shifting the feature channels of neighboring frames (as done in TSM); (2) Warping the shifted features to align with the current frame's features using optical flow; and (3) Combining the shifted feature maps with the original (unwarped) feature maps using the validity mask. Specifically, given feature map $X_{t}$, $X_{t-1}$ and $X_{t+1}$ at neighboring time stamps, we warp the first $f$ channels from $X_{t-1}$ to $X_{t}$ and the next $f$ channels from $X_{t+1}$ to $X_{t}$ using predicted flow:
\begin{equation}
\begin{split}
X_t'[0:f] = F_{(t-1) \to t}(X_{(t-1)}[0:f]), \\
X_t'[f:2f] = F_{(t+1) \to t}(X_{(t+1)}[f:2f])
\end{split}
\label{eq:shift}
\end{equation}
where $F_{* \to *}$ denotes the predicted optical flow and $X'_*$ denotes the aligned feature map. 

After obtaining the shift-and-aligned feature map, we combine the shifted and original (unwarped) features on the modified channels using the validity mask $v$:
\begin{equation}
X_t[0:2f] = v \; X_t'[0:2f] + (1-v) \; X_t[0:2f]
\end{equation}

The intuition is to borrow as much information as possible from neighboring frames as long as the optical flow is valid. After combining the feature maps from channels $0$ to $2f$, we concatenate all the feature maps along the channel dimension and pass it as input to the ensuing layer. We insert the TSAM module into every bottleneck block in the encoding stage of the network, and also insert it into the convolution layers in the decoding stage, as shown in Fig.~\ref{fig:pipeline}.

\label{objective}
\subsection{Loss Functions}
The loss functions of our model consist of a reconstruction loss $L_r$, perceptual loss $L_{p}$, style loss $L_{s}$, and temporal patchGAN~\cite{chang2019free} loss $L_G$:
\begin{equation}
\begin{split}
L_{total} = L_{r} +  
\lambda_{p}L_{p} + \lambda_{s}L_{s} + \lambda_{G}L_G    
\end{split}
\end{equation}
where $\lambda_{p}$, $\lambda_{s}$, $\lambda_{G}$ are coefficients for the loss terms.

\vspace{-5pt}
\paragraph{Reconstruction Loss.} Following most video inpainting works~\cite{chang2019free,chang2019learnable,zeng2020learning}, our reconstruction loss has two parts.  An L1 loss to constraint the overall reconstruction of the entire image and another L1 loss to focus the pixel reconstruction accuracy of the corrupted region:
\begin{equation}
    L_{r} = \lambda_a \sum_{t,i,j}{|Y_{t,i,j}' - Y_{t,i,j}|} + \\
    \lambda_c \sum_{(t,i,j) \in C}{|Y_{t,i,j}' - Y_{t,i,j}|}
\end{equation}
where $Y'$ and $Y$ are the predicted video and ground truth video, respectively. $C$ is the set of pixels in the corrupted region, and $\lambda_a$ and $\lambda_c$ are coefficients.

\vspace{-5pt}
\paragraph{Perceptual Loss.} Perceptual Loss \cite{johnson2016perceptual} is widely used in image or video inpainting tasks to improve the visual quality of the generated images:
\begin{equation}
L_{p} =  \sum_{t=1}^{n} \sum_{p \in P} \frac{ {||\phi_p^{Y'_t} - \phi_p^{Y_t}||}}{N_{ p }}     
\end{equation}
where $\phi_p^{Y'_t}$ and $\phi_p^{Y_t}$ denote the activation from the $p^{th}$ selected layer of a pre-trained network for the predicted ($Y'_t$) and ground truth ($Y_t$) video frame at time $t$, respectively. $N_{p}$ is the number of elements in the $p^{th}$ layer, $P$ is the set of layers used to compute the perceptual loss; specifically, we use the feature maps at the end of four convolutional stages from VGG network pre-trained on ImageNet.  The loss is accumulated over all frames in the generated video.

\begin{table*}[]
\footnotesize
\setlength{\tabcolsep}{2.5pt}
\caption{This table shows quantitative results on the FVI and DAVIS datasets. We compare our method with 6 different baselines under three different metrics using object mask, curve mask and stationary mask. Our model achieves state-of-the-art results.}
\vspace{4pt}
\resizebox{1\textwidth}{!}{
\begin{tabular}{c|ccc|ccc|ccc|ccc|ccc|ccc}
\hline
\multicolumn{1}{l|}{} & \multicolumn{9}{c|}{\textbf{FVI}}                                                                                                                           & \multicolumn{9}{c}{\textbf{DAVIS}}                                                                                                                          \\ \hline
                      & \multicolumn{3}{c|}{Object Mask}                  & \multicolumn{3}{c|}{Curve Mask}                    & \multicolumn{3}{c|}{Stationary Mask}               & \multicolumn{3}{c|}{Object Mask}                   & \multicolumn{3}{c|}{Curve Mask}                    & \multicolumn{3}{c}{Stationary Mask}                \\ \cline{2-19} 
                      & PSNR           & SSIM           & VFID            & PSNR           & SSIM            & VFID            & PSNR           & SSIM            & VFID            & PSNR           & SSIM            & VFID            & PSNR           & SSIM            & VFID            & PSNR           & SSIM            & VFID            \\ \hline
OPN \cite{oh2019onion}                & 33.53          & 0.8844         & 0.7618          & 34.16          & 0.9125          & 0.6602          & 36.15          & 0.9540          & 0.4004          & 32.91          & 0.8635          & 0.3664          & 33.78          & 0.9105          & 0.2701          & 36.33          & 0.9596          & 0.1281          \\
CPN \cite{lee2019copy}                  & 33.18          & 0.8764         & 0.8257          & 32.88          & 0.8676          & 0.8841          & 35.86          & 0.9485          & 0.4606          & 32.60          & 0.8452          & 0.4331          & 32.47          & 0.8496          & 0.4802          & 36.55          & 0.9547          & 0.1637          \\
FFVI \cite{chang2019free}                 & 34.74          & 0.8899         & 0.6946          & 36.84          & 0.9470          & 0.4099          & 35.23          & 0.9375          & 0.4543          & 33.45          & 0.8469          & 0.3809          & 35.76          & 0.9374          & \textbf{0.1843}          & 41.18          & 0.9679          & 0.1313          \\
DFGVI \cite{xu2019deep}                 & 33.33          & 0.8519         & 0.9122          & 32.22          & 0.8007          & 1.2020          & 37.46          & 0.9508          & 0.4838          & 32.78          & 0.8171          & 0.5169          & 32.02          & 0.7688          & 0.7337          & 38.20          & 0.9470          & 0.1894          \\
FGVC \cite{gao2020flow}                  & 33.13          & 0.8832         & 0.7640          & 34.14          & 0.9212          & 0.640           & 35.09          & 0.9422          & 0.4017          & 31.95          & 0.8323          & 0.4010          & 32.84          & 0.8841          & 0.3432          & 33.92          & 0.9212          & 0.1734          \\
STTN \cite{zeng2020learning}                  & 34.86          & 0.9047         & 0.7276          & 36.07          & 0.9411          & 0.6136          & 39.60          & 0.9716          & 0.3132          & 33.60          & 0.8708          & 0.3831          & 34.83          & 0.9251          & 0.2882          & 38.78          & 0.9690          & \textbf{0.1197} \\
Ours                  & \textbf{35.48} & \textbf{0.9160} & \textbf{0.6129} & \textbf{37.43} & \textbf{0.9566} & \textbf{0.3661} & \textbf{41.41} & \textbf{0.9738} & \textbf{0.2893} & \textbf{34.23} & \textbf{0.8798} & \textbf{0.3526} & \textbf{36.54} & \textbf{0.9508} & 0.1933 & \textbf{42.05} & \textbf{0.9737} & 0.1303          \\ \hline
\end{tabular}
}
\label{table:main}
\end{table*}

\vspace{-5pt}
\paragraph{Style Loss} We also apply the Style Loss~\cite{gatys2015neural}, which is also widely applied in image/video inpainting tasks.  It enforces that the predicted image and ground truth image have similar texture information (as measured by feature correlation).  The style loss is also accumulated over all frames between the generated video and the ground truth video.



\section{Experiments}
In this section, we first provide implementation details of our model. We then introduce the datasets used to evaluate our model, and provide training details for each dataset to reproduce our results. To demonstrate the effectiveness of our approach we compare our method with recent video inpainting methods both quantitatively and qualitatively. Finally, we ablate our model with various baseline components. 


\subsection{Implementation Details}

Our network structure is shown in Fig. \ref{fig:pipeline} and discussed in Section \ref{section:model}. We set $f$ in Fig. \ref{fig:fgtsm} to be $1/8$ of the total feature channels. During training we use the optical flow that is computed by \cite{ilg2017flownet} on ground truth image pairs. During evaluation, where ground truth flow is not available on the corrupted regions, we use FGVC  \cite{gao2020flow} to complete the optical flow within the missing regions. All flow maps are downsampled to the same resolution of the corresponding feature maps before feeding into the TSAM convolution.

\subsection{Datasets}

\label{FVI}
\paragraph{FVI \cite{chang2019free}}  samples data from the Youtube-VOS \cite{xu2018youtube} video object segmentation dataset . It contains 1940 videos for training, and 100 videos for testing. We use 100 additional videos that are not included in these videos for validation. The original FVI dataset also contains 12600 videos from the YTBB \cite{real2017youtube} dataset. However, we do not include them for training since~\cite{chang2019free} showed that they do not lead to any performance improvement.



\vspace{-5pt}
\paragraph{DAVIS \cite{Caelles_arXiv_2018}}  consists of 150 videos in total, in which 90 videos are densely annotated for training and 60 videos for validation are only annotated with the first frame. We follow \cite{zeng2020learning} and use the 60 original validation videos for training and the 90 original training videos for validation.



\begin{figure}[t]
\begin{center}
\includegraphics[width=.48\textwidth]{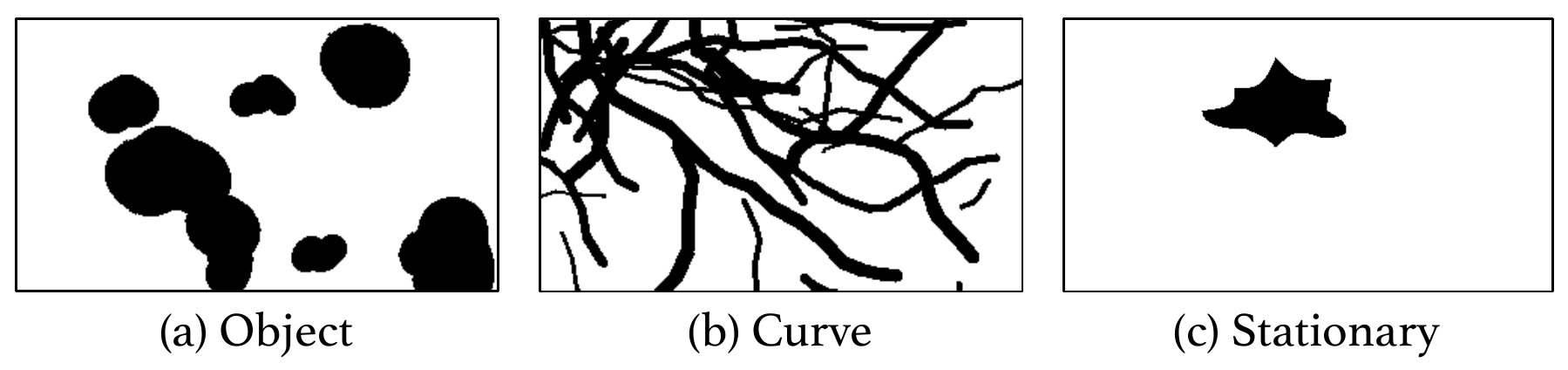}
\end{center}
\vspace{-10pt}
   \caption{Examples of the three types of corruption masks that we use for training and evaluation.}
\label{fig:mask}
\vspace{-15pt}

\end{figure}

\begin{table*}[]
\footnotesize
\setlength{\tabcolsep}{5.3pt}
\caption{This table shows ablation study results of our Temporal Feature Alignment Network. We first compare our full method (third row) with an ablated baseline that does not use optical flow (first row). We also compare to a baseline that uses ground truth optical flow  (second row). The result shows that using optical flow generally increases performance while using ground truth flow can further boost performance compared to using flow computed by \cite{gao2020flow}. Here Flow* denotes predicted flow and Mask* denotes the validity mask.}
\vspace{4pt}
\centering
\begin{tabular}{cccc|ccc|ccc|ccc}
\hline
\multicolumn{4}{l|}{} & \multicolumn{3}{c|}{Object Mask} & \multicolumn{3}{c|}{Curve Mask} & \multicolumn{3}{c}{Stationary Mask} \\ \hline
TSM  & Flow GT  & Flow*  & Mask* & PSNR     & SSIM      & VFID      & PSNR     & SSIM      & VFID     & PSNR       & SSIM       & VFID       \\ \hline
\checkmark    &          &        &       & 35.06    & 0.9047    & 0.6294    & 37.12    & 0.9510     & 0.3827   & 41.11      & 0.9694     & 0.2961     \\
\checkmark    & \checkmark        &        & \checkmark     & 35.76    & 0.9269    & 0.5404    & 37.82    & 0.9648    & 0.3660   & 41.61      & 0.9765     & 0.2549     \\
\checkmark    &          & \checkmark      & \checkmark     & 35.48    & 0.9160     & 0.6129    & 37.43    & 0.9566    & 0.3661   & 41.41      & 0.9738     & 0.2893     \\ \hline
\end{tabular}
\label{table:ablate}
\end{table*}

\vspace{-5pt}
\paragraph{Mask Types} Real-world applications of video inpainting include corrupted video restoration, object removal, watermark removal, etc. To mimic these applications, we train and evaluate our model on three kinds of masks including moving object-like mask, moving curve mask, and stationary mask as shown in Fig.~\ref{fig:mask}. The image regions within masks (black areas) are used as the corrupted regions in our task. The object-like/curve masks contain moving masks that occupy $0-10\%$ to $60-70\%$ of the overall frame area. 
Note that the object mask and curve mask are moving in both the evaluation and training stages. The stationary mask is static in evaluation but moving with a probability of 0.5 during training for data augmentation purposes.
The generation process of these masks follows previous work~\cite{zeng2020learning, chang2019free}.

\vspace{-5pt}
\paragraph{Training Details} On FVI and DAVIS datasets, we use slightly different training strategies. On the large-scale FVI dataset, we train our model in two stages. In the first stage, we only train the encoder and decoder network with image reconstruction loss, perceptual loss, and style loss with weights $1, 1, 2$ respectively. We train the network for 200 epochs. During the second stage, we add the image reconstruction loss in the corrupted region and temporal patch GAN loss to the loss function with weight 6 and 0.1 respectively. The model is further finetuned for 200 epochs.
As DAVIS only contains 60 videos for training, which is not sufficient to train the network from scratch, we use the model trained on FVI dataset as a pretrained model and finetune it on DAVIS. We train with reconstruction loss, style loss, perceptual loss, and temporal patch GAN loss, which use the same weights as on FVI. We finetune the model for 200 epochs on DAVIS. 

\subsection{Baselines and Evaluation Metrics}
\paragraph{Baselines} We compare our method with recent video inpainting algorithms including two attention-based methods OPN \cite{oh2019onion} and STTN \cite{zeng2020learning}, two flow-based approaches DFGVI \cite{xu2019deep} and FGVC \cite{gao2020flow}, one 3D convolution method FFVI~\cite{chang2019free}, and one affine alignment method CPN~\cite{lee2019copy}. 

\vspace{-5pt}
\paragraph{Evaluation Metrics}
We evaluate video inpainting quality using PSNR, SSIM and VFID.

Peak signal-to-noise ratio (PSNR) is a metric that measures the pixel similarity between the predicted and ground truth frames:
\begin{equation}
    PSNR = 20log_{10}\max(I) - 10 log_{10}(MSE)
\end{equation}
where $\max(I)$ is the maximum possible pixel value of the image, and MSE is the mean square error between the result and ground truth image. 

\begin{figure*}[t]
\begin{center}
\includegraphics[width=1.\textwidth]{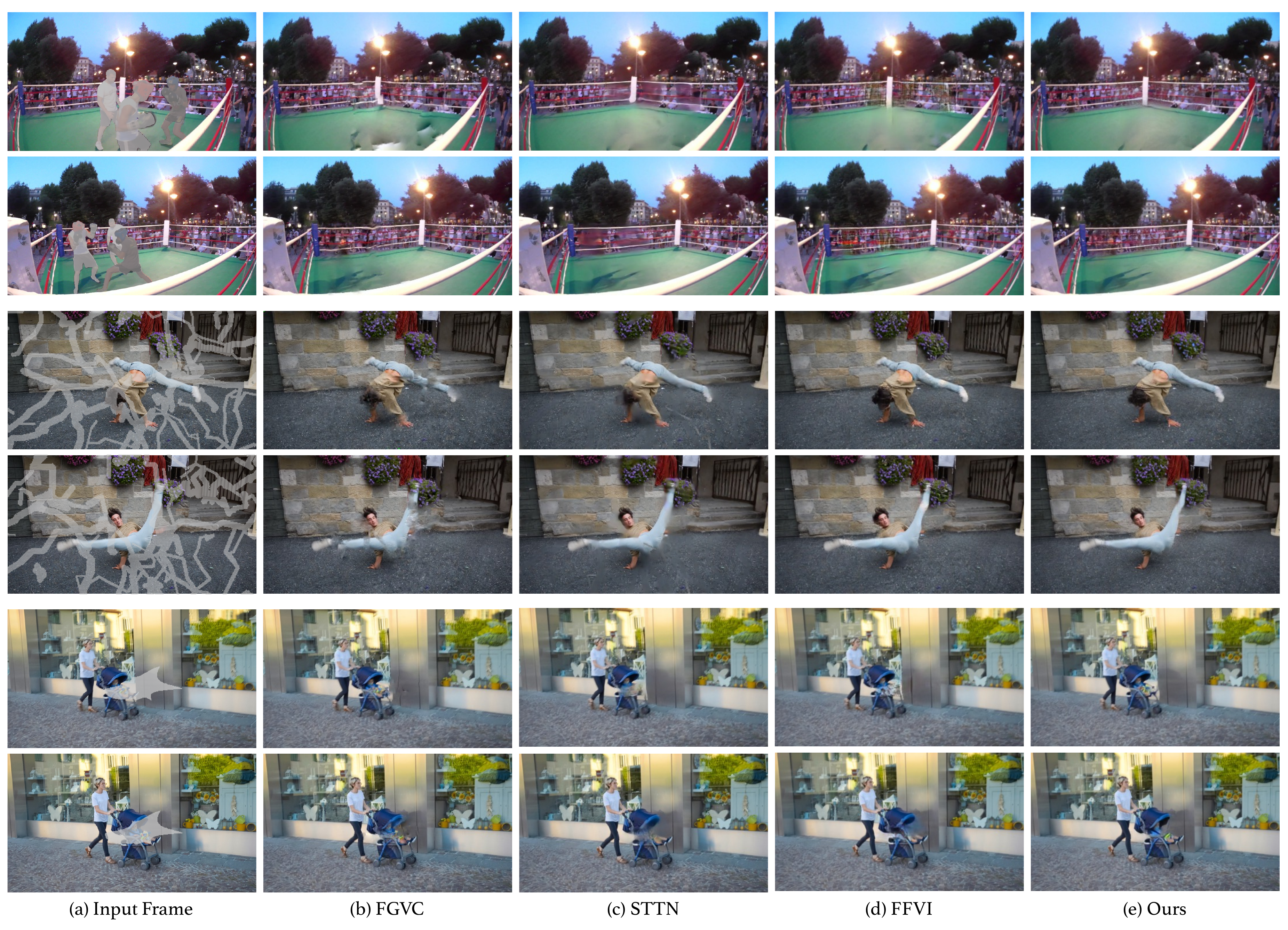}
\vspace{-12pt}
\end{center}
   \caption{Qualitative comparison of our method with FGVC \cite{gao2020flow}, STTN \cite{zeng2020learning} and FFVI \cite{chang2019free}. The boxing video shows that our approach has higher resolution on the inpainted area compared to STTN, and has better structure on the boxing fence compared to FGVC. The second video shows that our approach accurately inpaints the missing curve area. For the third example with a stationary mask, our approach fills the missing stroller area with plausible content.  In general, our approach fills the missing regions with more accurate content and higher resolution. }
\label{fig:qualitative}
\vspace{-8pt}

\end{figure*}

Structure Similarity (SSIM) measures the patch similarity between two images:
\begin{equation}
SSIM(p,q) =  \frac{(2\mu_p\mu_q +  \varepsilon_1 )(2 \sigma_{pq} + \varepsilon_2)}{(\mu_p^2 + \mu_q^2 + \varepsilon_1)(\sigma_{p}^2 + \sigma_{q}^2 + \varepsilon_2)} 
\end{equation}
where $\mu_p,\mu_q$ are the average of patch $p$ and $q$, $\sigma_{p},\sigma_q$ are the variance of patch $p$ and $q$, and $\sigma_{pq}$ is the covariance of patch $p$ and $q$. $\varepsilon_1, \varepsilon_2$ are two small constants to prevent division by 0.

Video Frechet Inception Distance (VFID) \cite{wang2018video} calculates the distance between features predicted by I3D \cite{carreira2017quo} models pretrained on an action recognition task:
\begin{equation}
FID = || \mu -  \mu' || + Tr(\Sigma  + \Sigma' - 2 \sqrt{\Sigma \Sigma'})
\end{equation}
where $\mu$ and $\Sigma$ are the mean and covariance of ground truth feature map and $\mu'$ and $\Sigma'$ are the mean and variance of the predicted feature map.


\subsection{Quantitative Results}
We report quantitative results on two datasets (DAVIS, FVI) and three different masks (object-like, curve, and stationary).

\vspace{-5pt}
\paragraph{FVI} Table \ref{table:main} (left) shows quantitative results on the FVI dataset. Our model outperforms all existing methods under the three different mask settings. In particular, our method outperforms FFVI, which uses 3D convolution to aggregate temporal features across frames but does not perform alignment. This indicates that the proposed temporal feature alignment is critical to improve the visual quality of the inpainted videos. Our method also outperforms FGVC with a large margin for all three evaluation metrics, especially on curve masks with a 42\% improvement. Although we use the same optical flow completion method as FGVC, our approach applies the inpainted flow at the feature level and uses the 3D convolution network to fill the hole in an end-to-end manner. STTN achieves the second best performance overall. STTN is an attention based approach that also iteratively fills the content through transformers. However, its attention modules are applied after feature encoding, whereas we progressively align feature maps during feature encoding across different feature scales. This enables the network to borrow both low and high level information from neighboring frames.

\vspace{-5pt}
\paragraph{DAVIS} Table \ref{table:main} (right) shows quantitative results on the DAVIS dataset. Overall, our model achieves the best performance on all three tasks. Compared to the previous best method STTN, our method produces better results on both object-like mask and curve mask. On stationary mask, our approach is significantly better on PSNR and SSIM, while being slightly worse on VFID score. We can see that flow-based methods have lower FID scores on curve masks than 3D convolution based approaches (\cite{chang2019free} and ours). This is because the curve mask is usually thin and moving, so there is sufficient surrounding (spatial and temporal) information for 3D convolution approaches to effectively hallucinate missing content, while FGVC \cite{gao2020flow} suffers when the predicted flow is inaccurate.



\subsection{Ablation Study}
\paragraph{With/out Optical Flow}
In Table \ref{table:ablate}, we first compare the video inpainting results using temporal shift module (first row) and our temporal shift-and-align module using ground truth optical flow (second row). Both approaches share exactly the same architecture shown in Fig.~\ref{fig:pipeline}. We see that using our temporal shift-and-align module with ground truth flow increases performance by around 5\% on PSRN, 4.6\% on SSIM, and 3.2\% on VFID. This demonstrates the importance of aligning features from neighboring frames.  

\vspace{-5pt}
\paragraph{Ground Truth Flow vs.~Completed Flow}
In Table~\ref{table:ablate} second and third rows, we compare our temporal shift-and-align module using ground truth flow versus optical flow completed by \cite{gao2020flow} for the corrupted regions. Although using completed flow decreases performance compared to using ground truth flow, it still leads to a performance gain compared to the baseline method of using temporal shift module without any optical flow alignment (first row in Table~\ref{table:ablate}).


\subsection{Qualitative Results}
Fig.~\ref{fig:qualitative} shows three sample video inpainting results for object removal, curve mask, and stationary mask corruption. We compare our approach with FGVC \cite{gao2020flow}, STTN \cite{zeng2020learning} and FFVI \cite{chang2019free}. In general, FGVC suffers from structure misalignment in all three cases (e.g. the boxing fence is misaligned, the man's legs are incorrectly inpainted, and the wall around the stroller also suffers from incorrect structure). Although STTN can inpaint plausible content, it tends to generate blurry results in most cases. Lastly, FFVI generates artifacts in some cases (e.g. yellow color around boxing fence, white blob on jeans). Compared to these methods, our approach generates more accurate content and structure in all three cases.


Finally, we also conduct a user study on the visual quality of different methods. Please refer to supplementary material for the results.

\vspace{-5pt}
\section{Conclusion}
In this paper, we proposed the progressive temporal feature alignment network for video inpainting, which fills the missing regions by making use of both temporal convolution and optical flow. We adopted the temporal shift module as our video backbone and used optical flow to align the features from neighboring frames on the shifted channels. This technique leads to inpainting results that have better image structure and higher resolution, which were limitations of prior approaches. We demonstrated state-of-the-art results on the FVI and DAVIS benchmark datasets, and the benefits of our novel model components through ablation studies.

\vspace{-10pt}
\paragraph{Acknowledgements.}  This work was supported in part by NSF IIS-1812850, AWS ML Research Award, Adobe Data Science Research Award. 

{\small
\bibliographystyle{ieee_fullname}
\bibliography{egbib}
}

\end{document}